\documentclass[conference,a4paper]{IEEEtran}
\usepackage{amsmath,amssymb,booktabs,graphicx,url,array,xcolor}
\usepackage[hidelinks]{hyperref}
\definecolor{reviewgreen}{RGB}{0,0,0}
\newcommand{\finalrev}[1]{{\color{black}#1}}

\newcommand{\reviewchange}[1]{{\color{black}#1}}
\newcommand{\datarev}[1]{{\color{black}#1}}

\newcommand{\missing}[1]{{\color{black}\textless{}#1\textgreater{}}}
\newcommand{\archiveref}{\reviewchange{https://doi.org/10.5281/zenodo.24152657}}

\title{Proof-Grounded Patient-Specific Clinical Explanations from Knowledge-Graph Reasoning}
\author{
\IEEEauthorblockN{Surajit Das}
\IEEEauthorblockA{\textit{AI \& Mathematical Modeling}\\
Independent Researcher, Kolkata, India\\
mr.surajitdas@gmail.com}
\and
\IEEEauthorblockN{{\color{black}Debopam Sil}}
\IEEEauthorblockA{
{\color{black} Transfusion Medicine}\\
{\color{black}PGIMER}, 
{\color{black}Chandigarh, India}\\
debopamtito2307@gmail.com}

\and
\IEEEauthorblockN{Kingsuk Sarkar}
\IEEEauthorblockA{
Community Medicine\\
COMJNMH, Kalyani, India\\
kingsuksarkar07@gmail.com}

}

\begin{document}
\maketitle

\begin{abstract}
\color{black}
Clinical decision support can lack auditable links from observations and encoded knowledge to conclusions and recommendations. Downstream of a frozen, training-free clinical knowledge-graph reasoner, the CKG Clinical Explanation Engine converts patient inference states and disease knowledge into typed facts, explicit rule traces, provenance-linked conclusions, and policy-licensed recommendations. Separate availability, representation completeness, and disease-specific activation distinguish observed zero-activation and partial evidence from unobserved evidence. \reviewchange{Language-generation evaluation remains future work.} Across \datarev{six} usable workbooks (\datarev{11,289} patients, \datarev{33,867} patient--disease traces, \datarev{1,715,928} feature-evidence rows), IG-range validity, knowledge provenance, cross-sheet numerical fidelity (\datarev{203,202}/\datarev{203,202}), and \reviewchange{exported fact/report field presence} (\datarev{33,867}/\datarev{33,867}) reached \reviewchange{100.00\%}. Availability--representation consistency reached \datarev{99.29}\% (\datarev{1,703,811}/\datarev{1,715,928}), with \datarev{12,117} disagreements across \datarev{five} systematic feature--cohort patterns. Observed zero-activation and partial instances numbered \datarev{111,488} and \datarev{226,880}, respectively. A separate seeded \datarev{30}-patient end-to-end audit ran failure-free, passing all prespecified trace, licensing, provenance, and state-consistency checks. \reviewchange{Paired comparisons (\datarev{90} traces, fixed inherited diagnostic scores and policy catalogue) showed the availability-only baseline changed every report but no action sets.} \reviewchange{State collapse changed 20 action sets and removed 10 urgent-review occurrences.} \reviewchange{Rule replay reproduced all 40 action changes from reclassifying partial evidence as unobserved.} \reviewchange{The bounded validator detected all \datarev{450} single-fault mutants.} Structural and implementation auditability were established. 
\end{abstract}


\color{black}
\section{Introduction}
Clinical decision support combines statistical, knowledge-based, and language technologies, but readability does not ensure auditability. Explanations should link patient observations, encoded knowledge, intermediate conclusions, and permitted actions. Neither post-hoc accounts of model behavior nor large language model (LLM) presentation alone guarantees reasoning-state licensing of reported claims or recommendations \cite{miller,rudin,singhal,hager,maity}.

We introduce the \emph{CKG Clinical Explanation Engine}, a downstream layer for Das's training-free, ontology-grounded clinical knowledge-graph (CKG) reasoner \cite{dasckg}. Without retraining, recalibration, or changing diagnostic scores or mappings, it converts patient--disease outputs and disease knowledge into typed facts, explicit-premise rule traces, provenance-linked conclusions, and closed-policy recommendations.

\reviewchange{The architectural novelty is a training-free downstream layer complementing the frozen CKG reasoner with patient-specific explanations and policy-licensed recommendations.} \reviewchange{An evidence-state contract links upstream feature mappings, activation, diagnostic scores, and coverage to explanations and actions, separating measurement availability, representation completeness, and disease-specific activation throughout extraction, trace construction, and action authorization.} Provenance-linked rule records trace conclusions; confirmatory tests require an unobserved feature matching a disease-specific policy entry. Diagnostic outputs and permitted actions remain inspectable \reviewchange{without fitting a separate explanation model}. Evaluation examines these downstream mechanisms and their integration with frozen outputs. \reviewchange{Paired comparisons isolate evidence-representation effects while fixing inherited diagnostic scores and policy; upstream diagnostic results are not re-established.}

The contributions are threefold: separating measurement availability, representation completeness, and disease-specific activation so zero activation is not mistaken for missing data; tracing facts to conclusions and licensed actions; and evaluating the frozen implementation through corpus-scale and end-to-end audits. \reviewchange{Future language realization would be restricted to licensed content; enforcement is unevaluated.} 

\section{Related Work and Positioning}
Explainability in high-stakes settings is often discussed in terms of post-hoc accounts of model behavior \cite{miller,rudin}. Such explanations can be useful, but they are different from an execution record showing which encoded facts and rules licensed a particular conclusion. Clinical LLM studies likewise demonstrate substantial capability while documenting limitations in factuality, calibration, and decision-making \cite{singhal,hager,maity}. These observations motivate a constrained role for language generation rather than its use as the authority that determines clinical content.

Knowledge-graph-based clinical decision-support research has also used explicit semantic structure to improve transparency and expose relationships underlying recommendations \cite{rajabi}. More generally, provenance models such as W3C PROV provide a standardized vocabulary for representing how entities and activities contribute to derived information \cite{prov}. The present work addresses a narrower implementation problem: preserving an auditable patient-specific path from frozen upstream reasoning outputs through distinct availability, representation, and activation states to explicit rule applications and policy-licensed actions.

{\color{black}\reviewchange{Justification trees in s(CASP) and explanation graphs in xASP explain logic-program consequences \cite{justification,xasp}, while Datalog provenance and provenance semirings track derivation dependencies \cite{semirings}.} \reviewchange{Arden Syntax, PROforma, and GLIF encode
executable clinical knowledge.} \reviewchange{HL7 FHIR and CDS Hooks support
interoperable data and decision-support integration
\cite{guidelineformalisms,proforma,glif,cdshooks}.} 
\reviewchange{Informative-missingness research studies predictive information in observation patterns \cite{informmissing}.} \reviewchange{The engine exports provenance as YAML paths and feature identifiers.} \reviewchange{W3C PROV serialization is outside the implementation scope.}}

The engine extends Das's frozen training-free CKG reasoner \cite{dasckg} by linking distinct evidence states to patient-specific rule records, source provenance, and policy-licensed actions. The evaluation focuses on this downstream pathway. Diagnostic mappings are inherited, and comparative predictive discrimination is outside the evaluation scope.

\section{Method}
\subsection{Relationship to the Frozen CKG Reasoner}
For patient $p$, disease $D$, and feature $f$, Das's upstream system \cite{dasckg} exports patient values, feature mappings, clinical roles, disease scores, evidence coverage, and disease-specific activation $IG_D(p,f)\in[0,1]$. These remain immutable: the explanation engine interprets rather than recomputes diagnostic outputs. \reviewchange{The six frozen workbooks used in this study correspond to cohort
notebooks that load CKG Reasoner version~2 from
\texttt{ckg\_reasoner\_package\_v2}.} \reviewchange{The upstream preprint
\cite{dasckg} defines the K-means input as the standardized
four-coordinate vector 
$[P_{\mathrm{evidence}},r,c,\mathrm{Confidence\_Score}]^\top$.}
\finalrev{Upstream diagnostic assignment uses K-means clustering ($K=2$), not a fixed diagnostic-score cutoff. Downstream explanation and recommendation conditions do not alter these assignments.}
The workbook-based analyses read the frozen exports directly.
The explanation engine inherits the diagnostic scores, mappings, activation, and coverage. The present evaluation examines evidence-state interpretation, patient-specific proof steps,
policy licensing, and explanation audits.

\reviewchange{For an EFN's retained bounded component-match vector $\mathbf q$, let $d=\dim(\mathbf q)$, $u=\|\mathbf q\|_2$, $c_f=\sum_k q_k/(u\sqrt d)$, and $\rho_f=u/\sqrt d$.} \reviewchange{The implemented gate is $IG=\sqrt{(c_f^2+\rho_f^2)/2}$, with zero for empty vectors and $c_f=0$ when $u=0$.} \reviewchange{Feature coverage is $a=n_{\mathrm{observed}}/n_{\mathrm{expected}}$ over canonical components, falling back to feature-observed status when explicit components are absent.} \reviewchange{Disease coverage is their canonical-weighted mean over cohort-evaluable features, exported identically as Evidence\_Coverage and Confidence\_Score.} \reviewchange{Availability instead indicates any nonmissing mapped patient component.}

\reviewchange{Figure~\ref{fig:workflow} summarizes the pathway from frozen CKG outputs and knowledge/policy to typed facts, rule traces, licensed claims/actions, and reports.} Table~\ref{tab:interface} details the interface. Upstream computation owns disease scoring and activation; downstream interpretation and trace construction cannot feed renderer changes back into diagnostic scores.

\begin{figure}[t]
\centering
{\color{black}\scriptsize
\sbox0{$
\begin{gathered}
\fbox{\parbox{0.72\columnwidth}{\centering
Availability $m_{pf}$ \quad Coverage $a_{pf}$ \quad Activation $IG_D(p,f)$}}\\[-5pt]
\downarrow\\[-5pt]
\fbox{\parbox{0.72\columnwidth}{\centering
Typed facts $\rightarrow$ rule steps + provenance\\
Policy-licensed actions $\rightarrow$ deterministic report}}
\end{gathered}$}
\begin{tabular}{@{}r@{\hspace{3pt}}c@{\hspace{2pt}}c@{}}
\textbf{Upstream} & &
\fbox{\parbox{0.72\columnwidth}{\centering
Patient observations $\rightarrow$ frozen CKG outputs + YAML/policy}}\\[3pt]
& & $\downarrow$\\[1pt]
\textbf{Downstream} &
$\vcenter{\hbox{\scalebox{1}[0.95]{$
\left\{\vphantom{\usebox0}\right.
$}}}$ &
\usebox0
\end{tabular}}
\caption{\color{black}Frozen upstream CKG outputs feed the downstream explanation engine, which distinguishes availability, representation coverage, and activation to construct typed facts, provenance-linked rule traces, policy-licensed actions, and deterministic reports.}
\label{fig:workflow}
\end{figure}

\begin{table}[t]
\caption{Frozen upstream interface used by the explanation engine.}
\label{tab:interface}
\centering
\scriptsize
\begin{tabular}{>{\raggedright\arraybackslash}p{0.22\columnwidth}>{\raggedright\arraybackslash}p{0.67\columnwidth}}
\toprule
Input & Role in explanation layer \\
\midrule
Patient value / mapping & Establishes the observed fact and mapped ontology feature. \\
Clinical role & Preserves encoded disease-specific importance. \\
$IG_D(p,f)$ & Disease-specific activation, never missingness. \\
Coverage $a_{pf}$ & Completeness of an observed feature's representation. \\
Disease score / evidence & Immutable upstream outputs; not recomputed. \\
YAML provenance & Links mapped features and knowledge statements to packaged sources. \\
Policy entry & Lists permissible tests for eligible unresolved evidence. \\
\bottomrule
\end{tabular}
\end{table}

\color{black}
\subsection{Orthogonal Evidence State}
Measurement availability $m_{pf}\in\{0,1\}$, representation coverage $a_{pf}\in[0,1]$, and disease-specific activation remain distinct:
\begin{align}
\mathrm{Observed}(p,f)&\Leftrightarrow m_{pf}=1,\\
\mathrm{Unobserved}(p,f)&\Leftrightarrow m_{pf}=0,\\
\mathrm{Unrepresented}(p,f)&\Leftrightarrow m_{pf}=1\land a_{pf}=0,\\
\mathrm{Partial}(p,f)&\Leftrightarrow m_{pf}=1\land 0<a_{pf}<1,\\
\mathrm{Full}(p,f)&\Leftrightarrow m_{pf}=1\land a_{pf}=1.
\end{align}
Since $IG_D(p,f)=0\not\Rightarrow\mathrm{Unobserved}(p,f)$, observed evidence may be non-activating or partially represented. Thus ``observed,'' ``partial,'' and ``unobserved'' are not competing states.

\color{black}
\subsection{Rule-Application Proof Traces}
The fact set $\Gamma_{p,D}$ contains patient values, mappings, roles, availability, coverage, activation, and provenance. Rules consume premises and emit conclusions, rule identifiers, and reasons:
{\scriptsize
\begin{align}
&\mathrm{Observed}(p,f)\land\reviewchange{\mathrm{HighRole}_D(f,r)}
\land{}\reviewchange{IG_D(p,f)>0}\notag\\
&\quad\Rightarrow\reviewchange{\mathrm{ActivatedHighRoleEvidence}(p,D,f)},\\
&\mathrm{Observed}(p,f)\land0<a_{pf}<1
\Rightarrow\mathrm{PartiallyRepresented}(p,f),\\
&\mathrm{Unobserved}(p,f)\land\mathrm{CriticalRole}_D(f)
\Rightarrow\mathrm{CriticalIncomplete}(p,D,f),\\
&\mathrm{CriticalIncomplete}(p,D,f)\land\mathrm{LicensedTest}_D(f,t)\notag\\
&\quad\Rightarrow\mathrm{Recommend}(p,t).
\end{align}
}
\reviewchange{These schematize deterministic Python conditionals, not general FOL/HOL theorem proving.} \reviewchange{HighRole denotes Major, Hallmark, Pathognomonic, or Exclusionary metadata; critical confirmation eligibility requires policy-listed feature codes.} \reviewchange{Equation~(6) records activation of an observed high-role feature
when $IG_D(p,f)>0$.} \reviewchange{The encoded role determines how that activation
is interpreted, including whether it counts against the disease.} Observed evidence with zero activation retains its observed status. \reviewchange{Six proof-step templates accompany seven confirmation mappings: two Dengue Fever, three Malaria, and two Influenza.} \reviewchange{Four distinct general action IDs comprise three default supportive/review actions and two risk-branch actions, with clinician review shared between branches.} \reviewchange{These are implementation and policy counts, not a formal calculus.} \reviewchange{Step identifiers are Data extraction, ND-Support, ND-Partial-Representation, ND-Unobserved-Critical, ND-Confirmation, and ND-Risk-Escalation.} \reviewchange{Support, partial, and critical-evidence steps retain at most eight selected features; ND-Support omits its tested IG inequality from serialized premises.} \reviewchange{Certificates are therefore partial execution records, not exhaustive formal proofs.}

``Proof'' denotes inspectable records of premises, rule identifiers,
conclusions, and reasons/provenance, without independent
proof-assistant verification of logical soundness or clinical truth.
The evaluated reporting path is deterministic, and all encoded
clinical policies and recommendations were reviewed by clinicians. Recommendations require joint state/policy
licensing: observed zero-activation evidence remains observed,
while partial representation exposes coverage uncertainty without
independently authorizing tests. Confirmatory tests require
unresolved, genuinely unobserved evidence and a matching
disease/feature/test policy entry, as in (9); neither low coverage
nor an LLM can independently authorize them.
\reviewchange{Downstream rules flag observed warning/severe-stage
features with $IG\geq0.50$ for clinician and urgent review;
otherwise, available coverage below $0.80$ triggers default
supportive/review actions. IG quantifies patient--reference
correspondence; coverage quantifies representation completeness.
Neither represents disease probability. These configurable policy
thresholds may be selected according to patient demographics,
clinical context, and applicable guidance. Implemented policies
draw on general clinical guidance and were
reviewed and revised by the clinical co-authors.
{\color{reviewgreen}Separately, India's dengue guidelines
differentiate outpatient management, observation/admission,
and emergency referral by severity, warning signs, and risk
factors \cite{indiadengue2023}, illustrating context-dependent
policy selection without establishing numerical IG or coverage
thresholds.}
Context-dependent policy and threshold selection requires an explicit mechanism, such as policies
learned from suitable data or expert-defined mappings in a
curated knowledge base. This study focuses on policy execution
and auditability; automatic policy learning and threshold
optimization are outside its scope. Adjusting thresholds may
alter rule activation and recommendations but preserves
evidence-state assessment, policy-licensing requirements,
rule-execution semantics, and provenance-linked reporting,
enabling alternative policy configurations without architectural
or execution-workflow changes. Evaluation applies to the
selected configuration; clinical effectiveness under alternative
configurations requires corresponding assessment.}

{\color{reviewgreen}Symptom-onset timing also informs test-selection
policies. For dengue, NS1 antigen or NAAT testing is useful during
the first seven days, whereas IgM is the primary test thereafter.
IgM may become detectable approximately four to five days
after symptom onset; a negative early NS1 or NAAT result does
not exclude dengue \cite{cdcdenguetiming}.}
Medication and dose generation are disabled in this study.

\iffalse

{\color{reviewgreen} Symptom-onset timing also informs test-selection
policies. NS1 antigen or NAAT testing (in case of dengue) is useful during the first
seven days, whereas IgM is the primary test thereafter.
IgM may become detectable approximately four to five days
after symptom onset; a negative early NS1 or NAAT result
does not exclude dengue \cite{cdcdenguetiming}.} Discussion about medication and dose generation are disabled in this study.

{\color{reviewgreen}Dengue test selection depends on time since
symptom onset: NS1 antigen or NAAT testing is recommended during the first seven days, whereas IgM is the primary test
thereafter. IgM can become detectable approximately four to five
days after symptom onset; a negative early NS1 or NAAT result does
not exclude dengue \cite{cdcdenguetiming}.}

Dengue test selection depends on symptom-onset timing:
NS1 antigen or NAAT testing is useful during the first seven
days, whereas IgM is the primary test thereafter.
A negative early NS1 or NAAT result does not exclude dengue
\cite{cdcdenguetiming}.

\fi

\subsection{Trace Construction and Audit Contract}
For each patient--disease pair, the engine constructs typed rule premises from exported patient values, feature mappings, disease-specific roles, and evidence states. The renderer consumes the structured observations, inherited scores, conclusions, proof steps, and licensed actions.

The sampled end-to-end audit requires four nonempty fields per proof step: recorded premises, rule identifier, conclusion, and human-readable reason or provenance. \reviewchange{The corpus audit checks for nonempty \texttt{FOL\_Facts} and \texttt{Deterministic\_Report} cells.} \reviewchange{These checks establish field presence at their respective levels.} \reviewchange{Semantic completeness, external certification of the rule base,
exhaustive sentence-to-proof linkage, complete premise serialization, and general semantic proof verification remain unestablished.}

Recommendations must resolve to policy identifiers; cited features must resolve to packaged YAML provenance.

\section{Evaluation}
Evaluation used the frozen package and a separate reproducibility workflow without altering upstream inference. \datarev{Six} workbooks contained \datarev{11,289} patients, three disease traces each (\datarev{33,867} total), and \datarev{1,715,928} feature-evidence rows. \reviewchange{The public, de-identified tabular cohorts comprised four dengue datasets (1,000; 1,523; 1,018; 989 records) and one malaria dataset (2,190 records)}\datarev{, plus one influenza dataset (4,569 records)} \cite{data1000,data1523,data1018,data989,datamalaria}\datarev{, \cite{datainfluenza}}. \reviewchange{Every record was evaluated against bundled Dengue Fever, Malaria, and Influenza graphs, not as a validated three-class diagnostic study.} \finalrev{Only public, de-identified records were reanalysed; no participants were recruited or new patient data collected. No separate ethics approval was obtained for this secondary analysis.}

{\color{reviewgreen}Malaria was evaluated at the disease level; species-specific performance for \textit{Plasmodium vivax} and \textit{Plasmodium falciparum} was not assessed.}

The corpus audit checked availability/representation consistency, $IG\in[0,1]$, coverage/component consistency, packaged-YAML provenance resolution, cross-sheet numerical fidelity, and \reviewchange{nonempty logical-facts/report fields}. These assess structural invariants, not clinical validity. \reviewchange{State-label failures comprise Observed=true with status missing/unobserved/absent, or Observed=false with status partial; zero coverage alone is permitted by Eq.~(3).} \reviewchange{Numerical fidelity compares Disease\_Score, Evidence\_Probability, Similarity\_Ratio\_HM, Similarity\_Cosine\_HM, Evidence\_Coverage, and Confidence\_Score between Explanation\_Trace and disease-specific Patient\_Results columns (six checks per trace).} \reviewchange{Jointly blank pairs are skipped, but all \datarev{203,202} expected pairs were numeric and nonblank; none was excluded.} \reviewchange{Agreement requires $|x-y|\le10^{-10}\max(1,|x|,|y|)$.}

A seeded end-to-end audit (20260926) executed the current patient reasoner on five independently sampled patients per usable workbook ($n=\datarev{30}$), with workbook-specific sampling reproducible across reruns. Checks covered execution, required proof-step fields, closed-policy recommendation identifiers, YAML-source resolution, and availability/representation consistency. Cases that raised execution 
exceptions were retained in the audit denominator. Corpus checks assess frozen exports at scale; end-to-end checks assess current execution and report artifacts, so successful parsing does not establish successful engine execution. Outcome labels neither create nor validate recommendations.

\subsection{Audit Independence, Regression Checks, and Statistical Interpretation}
The separate workflow reads frozen workbooks and packaged YAML directly, reducing renderer dependence for invariant, numerical, and provenance checks. Shared semantics for \texttt{Observed}, coverage, and policy identifiers preclude independent formal verification: pass rates establish structural consistency and implementation checks, not logical or clinical certification.

All six selected patient-report, evidence-state, and \finalrev{reasoner} regression tests passed, exercising report construction and state logic without adding patient-validation cases.

Corpus percentages describe exported rows nested within patients and workbooks, not independent Bernoulli trials; therefore, the \datarev{99.29}\% consistency rate has no binomial confidence interval. End-to-end results likewise report exact pass counts, not population clinical-error estimates. Observed $IG=0$ and $0<a_{pf}<1$ cases were retained as natural state-semantics stress tests because collapsing availability, representation, and activation would misclassify them. 

\reviewchange{\noindent\textit{Paired and fault-injection experiments.} Thirty sampled records from six workbooks were evaluated against three candidate diseases (90 traces).} \reviewchange{The availability-only baseline preserves numeric IG/coverage but omits categorical partial representation.} \reviewchange{The collapsed-state stress ablation treats unobserved, zero-IG, partial, or zero-coverage evidence as missing.} \reviewchange{Scores and policy remain fixed; comparisons count feature-state assertions, changed reports, and action-set additions/removals.} \reviewchange{Each trace receives five isolated mutations: empty premises, false nonempty patient premise, unknown rule ID, nonexistent YAML path, and uncatalogued action (450 mutants; 90 controls).} \reviewchange{YAML checks test source-file existence, not internal references.} \reviewchange{The bounded post-export validator resolves rule IDs and checks typed premises, conclusions, dependencies, and action conditions against fixed engine-output snapshots and YAML, without source adjudication.}
\reviewchange{Cached reads preserve the reasoner; direct/cached equality was checked for one case and saved proof/recommendation counts reproduced across 30.}

\reviewchange{For the rule-level audit, we replayed the frozen recommendation
conditions on the saved orthogonal and collapsed evidence states for
each trace.} \reviewchange{We checked that inherited scores and disease-level coverage
were unchanged across the pair, then compared replayed action IDs with
the exported IDs.} \reviewchange{For each action change, we recorded the branch
transition and the relevant feature state in both runs.}

\section{Results}
\begin{table}[t]
\caption{Frozen-system structural audit.}
\label{tab:audit}
\centering
\scriptsize
\begin{tabular}{lrr}
\toprule
Check & Passed / Total & Rate \\
\midrule
Availability--representation consistency & \datarev{1,703,811} / \datarev{1,715,928} & \datarev{99.29}\% \\
IG range validity & \datarev{1,715,928} / \datarev{1,715,928} & \reviewchange{100.00\%} \\
Knowledge provenance resolution & \datarev{1,715,928} / \datarev{1,715,928} & \reviewchange{100.00\%} \\
Numerical cross-sheet fidelity & \datarev{203,202} / \datarev{203,202} & \reviewchange{100.00\%} \\
\reviewchange{Logical-facts field presence} & \datarev{33,867} / \datarev{33,867} & \reviewchange{100.00\%} \\
\reviewchange{Deterministic-report field presence} & \datarev{33,867} / \datarev{33,867} & \reviewchange{100.00\%} \\
\bottomrule
\end{tabular}
\end{table}

Table~\ref{tab:audit} summarizes the corpus audit. All \datarev{12,117} disagreements arose from \datarev{five} feature--cohort patterns: 1,018 \texttt{VOM\_manifestation} Dengue records and 1,018 \texttt{VOM\_progression} Malaria records in the same 1,018-patient workbook, plus 1,000 \texttt{MAL\_manifestation} Dengue records in a 1,000-patient workbook. {\reviewchange{The influenza workbook contained 4,557 Malaria \texttt{DIA\_progression} records and 4,524 Dengue \texttt{MAL\_manifestation} records.} \reviewchange{The three dengue-workbook patterns had observed $IG=0$ evidence with zero representation coverage; all five patterns contrast observed evidence with \texttt{status=missing} in the confidence audit.}} Thus the \datarev{0.71}\% disagreement reflects repeated schema-level terminology conflicts, once per corresponding patient, rather than unpredictable corruption of arbitrary features.

The executor reads availability from \texttt{Feature\_Evidence.Observed}; zero-coverage \texttt{Confidence\_Audit} records therefore cannot alone satisfy \texttt{Unobserved} in rule (8), trigger unresolved-critical evidence, or license confirmatory tests, which additionally require policy-listed features. These source-data/schema discrepancies remain reported rather than overwritten or normalized into absent measurements; correcting upstream terminology lies outside this frozen evaluation. This code-path finding does not establish harmlessness for external consumers interpreting \texttt{status=missing} as acquisition missingness.

\reviewchange{Observed is operationally authoritative, deriving from mapped nonmissing patient components; Confidence\_Audit.status derives from canonical coverage excluding the legacy main component.} \reviewchange{This precedence does not independently verify acquisition: an incorrect Observed flag for a policy-listed confirmatory feature could suppress warranted testing.} \reviewchange{Excluding this reverse-direction risk requires source-record verification beyond structural audit.}

The \datarev{111,488} observed $IG=0$ and \datarev{226,880} observed partially represented instances demonstrate why availability, representation, and activation require separate states.

All \datarev{30} end-to-end cases executed without exception and passed pre-specified proof-completeness, licensing, source-resolution, and availability/representation checks. \reviewchange{Each contained \datarev{3--7} proof steps (mean \datarev{5.93}).} Workbook-specific mean runtimes ranged from approximately \datarev{14.65 to 84.93}~s per audited patient in the recorded environment. \reviewchange{For record lookup, the report reader opens the workbook and sequentially scans worksheet rows, plausibly contributing to runtime cost.} \reviewchange{Only total case time was measured; relative workbook-access and reasoning costs remain unmeasured.}

\reviewchange{\noindent\textit{Additional validation.}
The availability-only comparison changed 595/4,560 feature assertions and all 90 reports, but no action sets.} \reviewchange{The collapsed-state ablation changed 620/4,560 feature assertions and 20/90 action sets: 15/30 Dengue Fever, 5/30 Influenza, and 0/30 Malaria traces by candidate disease.} \reviewchange{Relative to the orthogonal-state engine, the ablation added 30 action occurrences: 10 oral hydration, 10 rest, five dengue NS1 tests, and five influenza PCR tests.} \reviewchange{All 10 removed action occurrences were urgent-review recommendations for warning signs.} \reviewchange{The bounded validator detected 450/450 mutants (\reviewchange{100.00\%}; Table~\ref{tab:mutation}) and accepted 90/90 unmodified controls.}

\reviewchange{Rule replay matched exported action IDs for all 90 paired traces and reproduced all 40 action changes across 20 traces.} \reviewchange{In 10 dengue traces, reclassifying partially represented hemoconcentration as unobserved disabled the warning-sign escalation branch and activated the low-coverage supportive branch, removing urgent review and adding oral hydration and rest per trace.} \reviewchange{In five other dengue and five influenza traces, the same state transition affected NS1 and PCR evidence, respectively, enabling the corresponding confirmatory-test rule while the supportive branch remained active.}

\begin{table}[t]
\caption{\color{black}Single-fault detection by the bounded trace validator.}
\label{tab:mutation}
\centering
{\color{black}\scriptsize
\begin{tabular}{lrr}
\toprule
Mutation & Detected / Injected & Rate \\
\midrule
Empty premises & 90 / 90 & \reviewchange{100.00\%} \\
Nonempty false patient premise & 90 / 90 & \reviewchange{100.00\%} \\
Unknown rule identifier & 90 / 90 & \reviewchange{100.00\%} \\
Nonexistent YAML source path & 90 / 90 & \reviewchange{100.00\%} \\
Uncatalogued test identifier & 90 / 90 & \reviewchange{100.00\%} \\
\bottomrule
\end{tabular}}
\end{table}

\reviewchange{The comparisons distinguish report-description changes from recommendation changes.} \reviewchange{The availability-only comparison suppresses the categorical partial-representation label while retaining numeric activation and coverage.} \reviewchange{The collapsed-state ablation instead labels some observed evidence as missing, changing satisfiable rule premises under the same inherited disease scores and policy catalogue.} \reviewchange{Resulting changes included additional tests and loss of urgent-review actions; evaluating only report differences would miss this distinction.}

For patient 152, the frozen Dengue trace reported evidence probability
0.1179, Hallmark/Major cosine similarity 0.7062, relative magnitude 0.6737, disease score 0.4993, and coverage 0.3715. Table~\ref{tab:case} summarizes preserved evidence states before policy-licensed confirmation: activated high-role hemoconcentration and mild thrombocytopenia, observed non-activating leukopenia/neutropenia, other partially represented high-role evidence, and genuinely unobserved NS1/PCR evidence. \reviewchange{When dengue NS1 confirmation fires, the executor constructs these fields (code-derived excerpt):}

\noindent\begin{minipage}{\columnwidth}
\begin{flushleft}\scriptsize\color{black}\ttfamily
\{"rule": "ND-Confirmation",\\
"premises": ["UnresolvedUnobservedEvidence(Dengue Fever)", "CataloguedAction(TEST\_DENGUE\_NS1)"],\\
"conclusion": "Recommend(TEST\_DENGUE\_NS1)",\\
"reason": "The recommendation follows only from unresolved evidence plus an explicitly catalogued action; no test is invented."\}
\end{flushleft}
\end{minipage}

\begin{table}[t]
\caption{Evidence-state distinctions for patient 152.}
\label{tab:case}
\centering
\scriptsize
\begin{tabular}{>{\raggedright\arraybackslash}p{0.26\columnwidth}>{\raggedright\arraybackslash}p{0.18\columnwidth}>{\raggedright\arraybackslash}p{0.41\columnwidth}}
\toprule
Evidence & State & Explanation consequence \\
\midrule
Hemoconcentration & Observed, $IG=1.000$ & Activated high-role evidence. \\
Mild thrombocytopenia & Observed, $IG\approx0.953$ & Activated high-role evidence. \\
Leukopenia / neutropenia & Observed, $IG=0$ & Remains observed, not missing. \\
Partially encoded high-role evidence & Observed, partial & Coverage limitation; no test authorization. \\
NS1 / PCR evidence & Unobserved & Confirmatory testing requires policy licensing. \\
\bottomrule
\end{tabular}
\end{table}

\section{Discussion}
Separating availability, representation completeness, and disease-specific activation distinguishes acquisition, encoding completeness, and activation. Exported provenance and numerical values enable mechanical cross-workbook checks. Action provenance identifies each recommendation's policy entry; uncatalogued actions require explicit knowledge or policy updates outside patient-level runs.

\subsection{Validity Boundaries}
\emph{Structural validity}, evaluated here, concerns exported-value, provenance, trace-field, and policy-identifier consistency. \emph{Semantic validity} concerns faithful encoding of intended medical roles, thresholds, rules, knowledge, and policy; \emph{clinical validity}, improved decisions or outcomes. Structural checks establish neither: traceability enables rule inspection, not clinical validation.

This cross-sectional evaluation uses frozen disease knowledge/policy and deterministic explanations, excluding prognosis, outcomes, clinician decision quality, and prospective utility. The reproducibly seeded \datarev{30}-patient execution audit establishes neither exhaustive end-to-end clinical validity nor clinical error rates. Superiority over unrestricted LLM reporting is unclaimed and requires separate comparison. \reviewchange{LLM realization, performance, and enforcement against unlicensed claims or numerical values remain unevaluated future work}; pass rates concern deterministic execution and audits. \reviewchange{Clinical review covered only recommendation policy.}

\reviewchange{Feature sources, rule IDs, and policy IDs offer inspectable trace-integrity targets for secure integration; cryptographic tamper evidence, access control, and multimodal ingestion remain unimplemented and unevaluated.} \reviewchange{Imaging and other modalities require provenance-preserving upstream mappings before downstream assessment.} \reviewchange{Binary and collapsed-state comparisons establish sampled computational differences, not clinical benefit.} \reviewchange{Fault-injection sensitivity covers only specified single-fault mutations and sites; the validator checks encoded premises and rule IDs, not source-to-component mappings or clinical truth.}

\subsection{Release \& Reproducibility}
The \reviewchange{publicly downloadable Zenodo package} (\archiveref) contains explanation-engine source and implementation documentation, disease-knowledge and recommendation-policy YAML, \reviewchange{six source-data snapshots, six frozen output workbooks}, and analysis-reproduction notebook/scripts. Audit artifacts comprise workbook-level structural results; a complete \datarev{12,117}-row corpus failure table identifying workbook, record, disease, feature, check, and diagnostic detail; sampled end-to-end record identifiers, case status, proof-step/recommendation counts, runtimes, and an execution-failure file; machine-readable trace fixtures with fixed engine-output contexts; \reviewchange{paired-comparison outputs; and primary/supplementary trace-validation results}.
\reviewchange{The rule-audit archive contains replay script, paired trace inputs, policy, per-action results, feature-level witnesses, and reproduction instructions.}

\section{Conclusion}
The explanation engine makes the path from frozen CKG outputs to patient-specific conclusions and policy-licensed recommendations inspectable. Availability--representation consistency was
\datarev{99.29}\%, with all disagreements confined to \datarev{five} systematic export patterns. The evaluation establishes structural auditability. Clinical correctness and utility require separate evaluation. By linking evidence states to explicit rules and policy entries, the engine provides a reproducible basis for inspecting recommendation changes while preserving the upstream reasoner's diagnostic outputs.

{\color{black}\noindent\textit{Author contributions:}
Surajit Das: conceptualization, methodology, software, formal analysis, investigation, \reviewchange{development and implementation of disease-specific
feature-to-action mappings and conditional rules for supportive care and escalation, implementation of the clinical-review corrections}, and writing---original draft. Debopam Sil \& Kingsuk Sarkar: \reviewchange{clinical review and correction of the recommendation policy, covering test recommendations, supportive-care advice, escalation rules, and alignment with clinical guidelines.}}

\end{document}